\documentclass[10pt,twocolumn,letterpaper]{article}

\usepackage{cvpr}
\usepackage{times}
\usepackage{epsfig}
\usepackage{graphicx}
\usepackage{amsmath}
\usepackage{amssymb}
\usepackage{multirow}
\usepackage{array}
\usepackage{booktabs}
\usepackage{comment}
\usepackage{subcaption}
\usepackage{threeparttable}
\usepackage{enumitem}
\usepackage{color}
\usepackage{url}

\newcommand{\ieno}{\textit{i}.\textit{e}.}
\newcommand{\egno}{\textit{e}.\textit{g}.} %there is no space

\usepackage[breaklinks=true,bookmarks=false]{hyperref}

\cvprfinalcopy % *** Uncomment this line for the final submission

\def\cvprPaperID{****} % *** Enter the CVPR Paper ID here

\ifcvprfinal\fi
\begin{document}

%%%%%%%%% TITLE
\title{Semantics-Guided Neural Networks for Efficient Skeleton-Based \\
Human Action Recognition}

\author{{Pengfei Zhang{\small $~^{1}$}\thanks{This work was done when P. Zhang was an intern at MSRA.}}, ~Cuiling Lan{\small $~^{2}$}\thanks{Corresponding author.},~  Wenjun Zeng{\small $~^{2}$}, ~Junliang Xing{\small $~^{3}$},   ~Jianru Xue{\small $~^{1}$}, ~Nanning Zheng{\small $~^{1}$}\\
	\normalsize
	$^{1}$\	Xi'an Jiaotong University, Shaanxi, China ~~ $^{2}$\,Microsoft Research Asia, Beijing, China\\
	\normalsize
	$^{3}$\,National Laboratory of Pattern Recognition, Institute of Automation, Chinese Academy of Sciences, Beijing, China \\
	\normalsize
	zpengfei@stu.xjtu.edu.cn,
	\{culan,wezeng\}@microsoft.com, 
	%\normalsize
	jlxing@nlpr.ia.ac.cn,  
	\{jrxue,nnzheng\}@mail.xjtu.edu.cn
	}
\maketitle
\thispagestyle{empty}

%%%%%%%%% ABSTRACT
\begin{abstract}
Skeleton-based human action recognition has attracted great interest thanks to the easy accessibility of the human skeleton data. Recently, there is a trend of using very deep feedforward neural networks to model the 3D coordinates of joints without considering the computational efficiency. In this paper, we propose a simple yet effective semantics-guided neural network (SGN) for skeleton-based action recognition. We explicitly introduce the high level semantics of joints (joint type and frame index) into the network to enhance the feature representation capability. In addition, we exploit the relationship of joints hierarchically through two modules, \ieno, a joint-level module for modeling the correlations of joints in the same frame and a frame-level module for modeling the dependencies of frames by taking the joints in the same frame as a whole. A strong baseline is proposed to facilitate the study of this field. With an order of magnitude smaller model size than most previous works, SGN achieves the state-of-the-art performance on the NTU60, NTU120, and SYSU datasets. The source code is available at \url{https://github.com/microsoft/SGN}.

\end{abstract}

\section{Introduction}

\begin{figure}[!t]
	\begin{center}
		\includegraphics[width=0.99\linewidth]{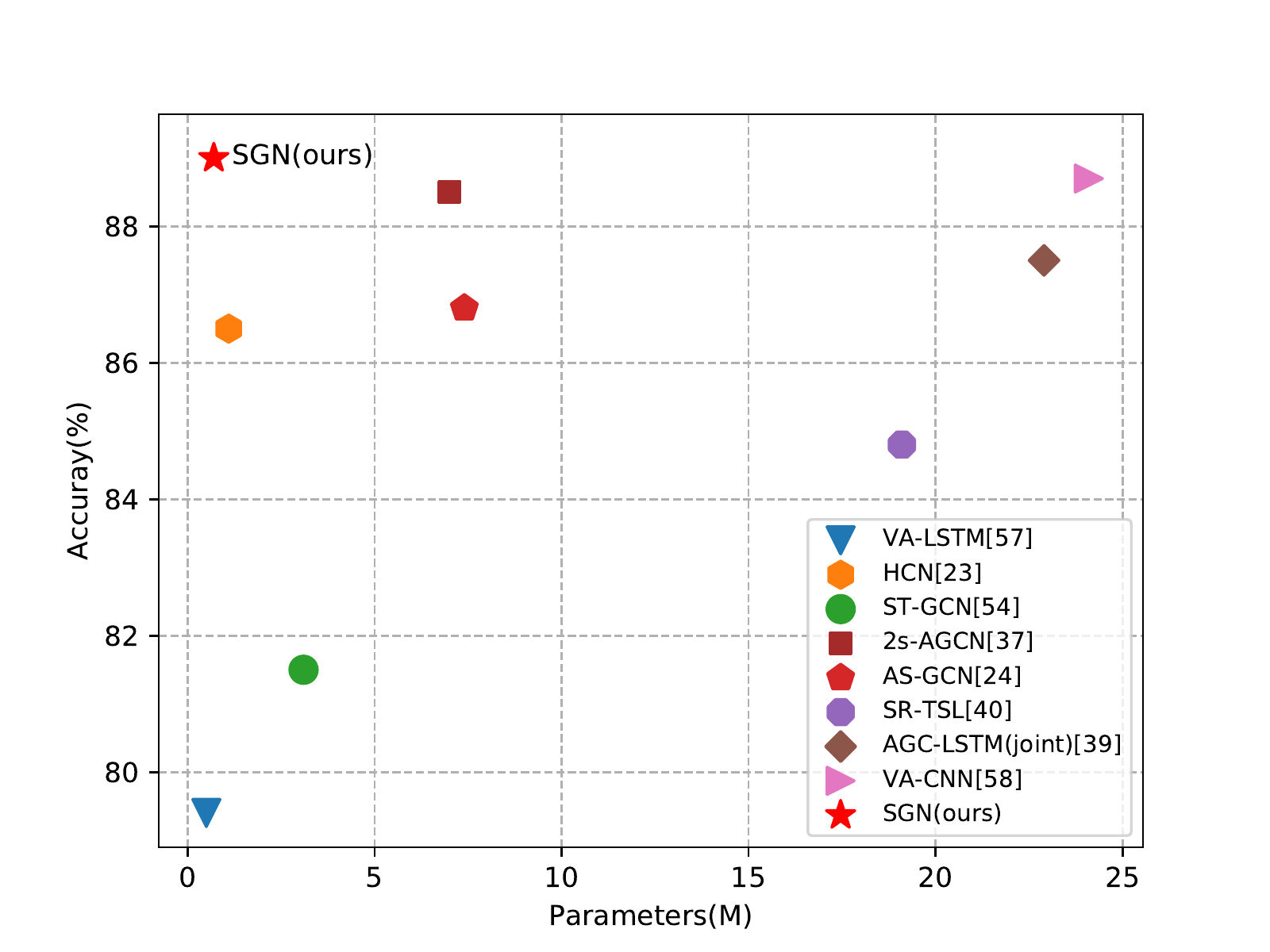}
	\end{center}
	\vspace{-4mm}
	\caption{Comparisons of different methods on NTU60 (CS setting) in terms of accuracy and the number of parameters. The proposed SGN model achieves the best performance with an order of magnitude smaller model size.}
	\label{fig:paras}
\end{figure}

Human action recognition has a wide range of application scenarios, such as human-computer interaction and video retrieval \cite{poppe2010survey, weinland2011survey, aggarwal2011human}. In recent years, skeleton-based action recognition \cite{yun2012two, du2015hierarchical, shahroudy2016ntu, zhang2019view} is attracting increasing interests. Skeleton is a type of well structured data with each joint of the human body identified by a joint type, a frame index, and a 3D position. There are several advantages of using the skeleton for action recognition. First, skeleton is a high level representation of the human body with the human pose and motion abstracted. Biologically, human is able to recognize the action category by observing only the motion of joints even without appearance information \cite{johansson1973visual}. Second, the advance of cost effective depth cameras \cite{zhang2012microsoft} and pose estimation technology \cite{shotton2011real,cao2017realtime,sun2019deep} make the access of skeleton much easier. Third, compared with RGB video, the skeleton representation is robust to variation of viewpoint and appearance. Fourth, it is also computationally efficient because of low dimensional representation. Besides, skeleton-based action recognition is also complementary to the RGB-based action recognition \cite{song2018skeleton}. In this work, we focus on skeleton-based action recognition.

\begin{figure*}[!t]
	\begin{center}
		\includegraphics[width=1\linewidth]{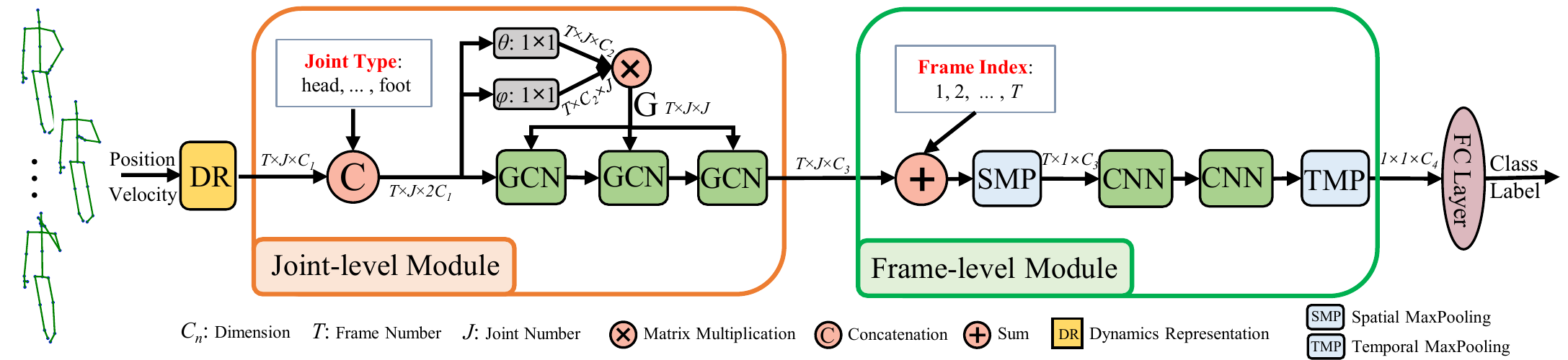}
	\end{center}
	\caption{Framework of the proposed end-to-end Semantics-Guided Neural Network (SGN). It consists of a joint-level module and a frame-level module. In DR, we learn the dynamics representation of a joint by fusing the position and velocity information of a joint. Two types of semantics, \ie, joint type and frame index, are incorporated into the joint-level module and the frame-level module, respectively. To model the dependencies of joints in the joint-level module, we use three GCN layers. To model the dependencies of frames, we use two CNN layers.}
	\label{fig:framework}
\end{figure*}

For skeleton-based action recognition, deep learning is widely used to model the spatio-temporal evolution of the skeleton sequence \cite{han2017space,wang2018rgb}. Various network structures have been exploited, such as Recurrent Neural Networks (RNN) \cite{du2015hierarchical, zhu2016co, shahroudy2016ntu, song2017end, zhang2017view, si2018skeleton}, Convolutional Neural Networks (CNN) \cite{ke2017new, zhang2019view, liu2017enhanced, weng2018deformable}, and Graph Convolutional Networks (GCN) \cite{yan2018spatial, si2018skeleton, tang2018deep}. In the early years, RNN/LSTM was the favored network to be used to exploit the short and long term temporal dynamics. Recently, there is a trend of using feedforward ({\it{i.e.}}, non-recurrent) convolutional neural networks for modeling sequences in speech, language \cite{oord2016wavenet,gehring2017convolutional,xiong2018microsoft,wang2018non}, and skeleton \cite{ke2017new,zhang2019view, liu2017enhanced, weng2018deformable} due to their superior performance. Most skeleton-based approaches organize the coordinates of joints to a 2D map and resize the map to a size (\eg 224$\times$224) suitable for the input of a CNN (\eg ResNet50~\cite{he2016deep}). Its rows/columns correspond to the different types of joints/frames indexes. In these methods \cite{ke2017new,zhang2019view, liu2017enhanced, weng2018deformable}, long-term dependencies and semantic information are expected to be captured by the large receptive fields of deep networks. This appears to be brutal and typically results in high model complexity. 
 
Intuitively, semantic information, {\it{i.e.}}, the joint type and the frame index, is very important for action recognition. Semantics together with dynamics ({\it{i.e.}}, 3D coordinates) reveal the spatial and temporal configuration/structure of human body joints. As we know, two joints of the same coordinates but different semantics would deliver very different information. For example, for a joint above the head, if this joint is a hand joint, the action is likely to be \emph{raising hand}; if it is a foot joint, the action may be \emph{kicking a leg}. Besides, the temporal information is also important for action recognition. Taking the two actions of \emph{sitting down} and \emph{standing up} as examples, they are different only in occurrence order of the frames. However, most approaches \cite{han2017space,wang2018rgb} overlook the importance of the semantic information and under-explore it.

To address the above mentioned limitations of current approaches, we propose a semantics-guided neural network (SGN) which explicitly exploits the semantics and dynamics for high efficient skeleton-based action recognition. Fig.~\ref{fig:framework} shows the overall framework. We build a hierarchical network by sequentially exploring the joint-level and frame-level dependencies of the skeleton sequence. For better joint-level correlation modeling, besides the dynamics, we incorporate the semantics of joint type (\egno, `head', and `hip') to the GCN layers which enables the content adaptive graph construction and effective message passing among joints within each frame. For better frame-level correlation modeling, we incorporate the semantics of temporal frame index to the network. Particularly, we perform a Spatial MaxPooling (SMP) operation over all the features of the joints within the same frame to obtain frame-level feature representation. Combined with the embedded frame index information, two temporal convolutional neural network layers are used to learn feature representations for classification. In addition, we develop a strong baseline which is of high performance and efficiency. Thanks to the efficient exploration of semantic information, the hierarchical modeling, and the strong baseline, our proposed SGN achieves the state-of-the-art performance with a much smaller number of parameters. 
%To model the joint-level dependencies, we first fuse the semantics of joint type to make the network realize the types of different joints, such as `left hand', `right leg', and `head'. The semantics-aware input will be used for constructing the content adaptive graph and message passing in the GCN layer. In the joint-level module, GCN is only applied to model the dependencies of joints in the same frame. To model the frame-level dependencies of joints, we first incorporates the semantics of temporal index to make the network realize the order of different frames, such as `first frame' and `last frame'. A Spatial MaxPooling (SMP) layer is then used to fuse all joints information within the same frame. The semantics-aware information is then fed into a temporal convolutional layer which models the temporal dependencies. Another CNN layer (with kernel size as 1) then maps the input to a high dimension space. Finally a Temporal MaxPooling (TMP) layer fuses all the information across frames to have a sequence-level feature vector for classification. In addition, we develop a strong baseline which is of high performance and efficiency. Thanks to the efficient exploration of semantic information, the hierarchical modeling, and the strong baseline, our proposed SGN achieves the state-of-the-art performance with a much smaller number of parameters.

% One CNN layer with kernel size as 1 then maps the input to a high dimension space. 

We summarize our three main contributions as follows:

\begin{itemize}[leftmargin=*,noitemsep,nolistsep]
%\begin{itemize}
	\item  We propose to explicitly explore the joint semantics (frame index and joint type) for efficient skeleton-based action recognition. Previous works overlook the importance of semantics and rely on deep networks with high complexity for action recognition.   
	\item We present a semantics-guided neural network (SGN) to exploit the spatial and temporal correlations at joint-level and frame-level hierarchically. 
	\item We develop a lightweight strong baseline, which is more powerful than most previous methods. We hope the strong baseline will be helpful for the study of skeleton-based action recognition.
	% with a small number of parameters
	% by exploiting recent technical skills
\end{itemize}

With the above technical contributions, we have obtained a high performance skeleton-based action recognition model with high computational efficiency. Extensive ablation studies demonstrate the effectiveness of the proposed model design. On the three largest benchmark datasets for skeleton-based action recognition, the proposed model consistently achieves superior performances over many competing algorithms while having an order of magnitude smaller model size than many algorithms (see Fig.~\ref{fig:paras}).

\section{Related Work}

Skeleton-based action recognition has attracted increasing attentions recently. Recent works using neural networks \cite{han2017space} have significantly outperformed traditional approaches that use hand-crafted features \cite{han2017space,xia2012view, wang2012mining, yu2014discriminative, garcia2017transition}. 

\noindent\textbf{Recurrent Neural Network based.} Recurrent neural networks, such as LSTM~\cite{hochreiter1997long} and GRU \cite{cho2014learning}, are often used to model the temporal dynamics of skeleton sequence \cite{du2015hierarchical, shahroudy2016ntu, zhu2016co,song2017end, zhang2017view, zhang2018adding, zhang2019eleatt}. The 3D coordinates of all joints in a frame are concatenated in some order to be the input vector of a time slot. They do not explicitly tell the networks which dimensions belong to which joint. Some other RNN-based works tend to design special structures in RNN to make it aware of the spatial structural information. Shahroudy {\it et al.} divide the cell of LSTM  into five sub cells corresponding to five body parts, \ie, torso, two arms, and two legs, respectively \cite{shahroudy2016ntu}. Liu {\it et al.} propose a spatial-temporal LSTM model to exploit the contextual dependency of joints in both the temporal and spatial domain \cite{liu2016spatio}, where they feed different types of joints at each step. To some extent, they distinguish the different joints.    

\noindent\textbf{Convolutional Neural Network based}. In recent years, in the field of speech, language sequence modeling, convolutional neural networks demonstrate their superiority in both accuracy and parallelism \cite{oord2016wavenet,gehring2017convolutional,xiong2018microsoft,wang2018non, vaswani2017attention}. The same is true for skeleton-based action recognition \cite{du2015skeleton, li2017skeleton, ke2017new,cao2018skeleton}. These CNN-based works transform the skeleton sequence to skeleton map of some target size and then use a popular network, such as ResNet \cite{he2016deep}, to explore the spatial and temporal dynamics. Some works transform a skeleton sequence to an image by treating the joint coordinate ({\emph{x,y,z}}) as the R, G, and B channels of a pixel \cite{du2015skeleton, li2017skeleton}. Ke {\it et al.} transform the skeleton sequence to four 2D arrays, which are represented by the relative position between four selected reference joints (\ie, the left/right shoulder, the left/right hip) and other joints \cite{ke2017new}. Skeleton is well structured data with explicit high level semantics, \ie, frame index and joint type. However, the kernels/filters of CNNs are translation invariant \cite{long2015fully} and thus cannot directly perceive the semantics from such input skeleton maps. The CNNs are expected to be aware of such semantics through large receptive fields of deep networks, which is not very efficient.

\noindent\textbf{Graph Convolutional Network based.} Graph convolutional networks \cite{kipf2016semi}, which have been proven to be effective for processing structured data, have also been used to model the structured skeleton data. Yan {\it et al.} propose a spatial and temporal graph convolutional network \cite{yan2018spatial}. They treat each joint as a node of the graph. The presence of edge denoting the joint relationship is pre-defined by human based on prior knowledge. To enhance the predefined graph, Tang {\it et al.} define the edges for both physically disconnected and connected joint pairs for better constructing the graph \cite{tang2018deep}. A SR-TSL model \cite{si2018skeleton} is proposed to learn the graph edge of five human body parts within each frame using a data-driven method instead of leveraging human definition. A two-stream GCN model \cite{shi2019two} learns a content adaptive graph based on the non-local block and uses it to pass messages in GCN layers. However, the informative semantics is not utilized for learning the graph edge and message passing of GCN, which makes the network less efficient. 

% In summary, previous works for skeleton-based action recognition overlook the importance of explicitly exploiting the semantics, \ie, frame index and joint type. In this work, we investigate the explicit exploitation of semantics leveraging both GCN and CNN and design a simple yet effective neural network for skeleton-based action recognition.

\noindent\textbf{Explicit Exploration of Semantics Information.} The explicit exploration of semantics has been exploited in other fields, \eg, machine translation \cite{vaswani2017attention} and image recognition \cite{zheng2019learning}. Ashish {\it{et al.}} explicitly encode the position of the tokens in the sequence to make use of the order of the sequence in machine translation tasks \cite{vaswani2017attention}. Zheng {\it{et al.}} encode the group index into convolutional channel representation to preserve the information of group order \cite{zheng2019learning}. For skeleton-based action recognition, however, the joint type and frame index semantics are overlooked even though such information is very important. In our work, we propose to explicitly encode the joint type and frame index to preserve the important information of the spatial and temporal body structure. As an initial attempt to explore such semantics, we hope it will inspire more investigation and exploration in the community.
%{\color{blue}\noindent\textbf{The explicit modeling of semantic information} has been exploited in other fields, \eg, machine translation \cite{vaswani2017attention} and image recognition \cite{zheng2019learning}. Ashish {\it{et al.}} \cite{vaswani2017attention} explicitly encode the position of the tokens in the sequence to make use of the order of the sequence in machine translation tasks. Zheng {\it{et al.}} \cite{zheng2019learning} encode the group index into convolutional channel representation to preserve the important information of group order. For skeleton-based action recognition, however, the joint type and frame index semantics are overlooked even though such information is very important. In our work, we propose to explicitly encode the joint type and frame index to preserve the important information of the spatial and temporal body structure. As an initial attempt to explore such semantics, we hope it will inspire more investigation and exploration in the community.}

\section{Semantics-Guided Neural Networks}

For a skeleton sequence, we identify a joint by its semantics (joint type and frame index) and represent it together with its dynamics (position/3D coordinates and velocity). Without semantics, the skeleton data will lose the important spatial and temporal structure. Previous CNN-based works \cite{ke2017new, du2015skeleton, zhang2019view}, however, typically overlook the semantics by implicitly hiding them in the 2D skeleton map (\eg with rows corresponding to the different types of joints and columns corresponding to the frame indexes). 

% We propose a semantics-guided neural network (SGN) for skeleton-based action recognition and show the overall end-to-end framework in Fig.~\ref{fig:framework}. For a skeleton sequence, we first identify a joint in DR by combining position and velocity information together. Then, a joint-level module exploits the correlations of joints in the same frame under the guidance of the semantics of joint type. Afterward, a frame-level module exploits the correlations of frames by taking all joints in the same frame as a whole under the guidance of the semantics of temporal index. More details about each component are described below.

We propose a semantics-guided neural network (SGN) for skeleton-based action recognition and show the overall end-to-end framework in Fig.~\ref{fig:framework}. It consists of a joint-level module and a frame-level module. We describe the details of the framework in the following subsections.

Specifically, for a skeleton sequence, we denote all the joints as a set $\mathcal{S}$ = \{$X_t^k$ $\mid$ $t = 1, 2, \dots, T; k= 1, 2, \dots, J$\}, where $X_t^k$ denotes the joint of type $k$ at time $t$. $T$ denotes the number of frames of the skeleton sequence and $J$ denotes the total number of joints of a human body in a frame. For a given joint $X_t^k$ of type $k$ at time $t$, it can be identified by its dynamics and semantics. Dynamics are related to the 3D position of a joint. Semantics means the frame index $t$ and joint type $k$.

\subsection{Dynamics Representation}

For a given joint $X_t^k$, we define its dynamics by the position $\mathbf{p}_{t,k} = (x_{t,k}, y_{t,k}, z_{t,k})^T \in \mathbb{R}^3$ in the 3D coordinate system, and the velocity $\mathbf{v}_{t,k} = \mathbf{p}_{t,k} - \mathbf{p}_{t-1,k}$. We encode/embed the position and velocity into the same high dimensional space, \ie, $\widetilde{\mathbf{p}_{t,k}}$ and $\widetilde{\mathbf{v}_{t,k}}$, respectively, and fuse them together by summation as
\begin{equation}
\textbf{z}_{t,k} = \widetilde{\mathbf{p}_{t,k}}  +  \widetilde{\mathbf{v}_{t,k}} \in \mathbb{R}^{C_{1}},
\end{equation}
\label{eq:fuse}
where $C_{1}$ is the dimension of the joint representation.
Take the embedding of position as an example, we encode the position $\mathbf{p}_{t,k}$ using two fully connected (FC) layers as 
\begin{equation}
    \widetilde{\mathbf{p}_{t,k}} = \sigma(W_2(\sigma(W_1\mathbf{p}_{t,k} + \mathbf{b}_1))+ \mathbf{b}_2),
    \label{equ:mapping}
\end{equation}
where $W_1 \in \mathbb{R}^{C_1 \times 3}$ and $W_2 \in \mathbb{R}^{C_1 \times C_1}$ are weight matrices, $\mathbf{b}_1$ and $\mathbf{b}_2$ are the bias vectors, $\sigma$ denotes the ReLU activation function \cite{nair2010rectified}. Similarly, we obtain the embedding for velocity as $\widetilde{\mathbf{v}_{t,k}}$.

%Two pieces of information describe a joint from different aspects. We aggregate them together through sum to fully represent one joint as dynamics $\textbf{z}_{t,j} = \widetilde{\mathbf{p}_{t,j}} $ + $ \widetilde{\mathbf{v}_{t,j}} \in \mathbb{R}^{C_{1}}$, where $C_{1}$ is the dimension of the joint representation.
 
%Two pieces of information are encoded to a high dimensional space and fused by sum. Because the two pieces of dynamics are not in the same domain, separate embedding in the same manner is performed with unshared parameters.

\subsection{Joint-level Module}
We design a joint-level module to exploit the correlations of joints in the same frame. We adopt graph convolutional networks (GCN) to explore the correlations for the structural skeleton data. Some previous GCN-based approaches take the joints as nodes and they pre-define the graph connections (edges) based on prior knowledge \cite{yan2018spatial}  or learn a content adaptive graph \cite{shi2019two}. We also learn a content adaptive graph, but differently we incorporate the semantics of joint type to the GCN layers for more effective learning.%, which makes it inefficient.
%to effectively help the graph construction and let the semantics take part in the message passing

We enhance the power of GCN layers by making full use of the semantics from two aspects. First, we use the semantics of joint type and the dynamics to learn the graph connections among the nodes (different joints) within a frame. The joint type information is helpful for learning suitable adjacent matrix (\ieno, relations between joints in terms of connecting weights). Take two source joints, \emph{foot} and \emph{hand}, and a target joint \emph{head} as an example, intuitively, the connection weight value from \emph{foot} to \emph{head} should be different from the value from \emph{hand} to \emph{head} even when the dynamics of \emph{foot} and \emph{hand} are the same. Second, as part of the information of a joint, the semantics of joint types takes part in the message passing process in GCN layers. 
%The GCN is position-independent and it has no idea of the structural information of skeleton, which is harmful for recognizing actions. We thus take advantage of the semantics of joint type to make the GCN know the structural information during massage passing. 

%We use the semantics of joint type to guide GCN. First, the semantics of joint type will be used to learn graph connections of different joints in the same frame. As is known, the dynamics of different joint pairs may be the same. However, the connected edge weights should be different since the relationships are different for pairs of different semantics. We thus propose to use both dynamics and semantics of joint type to learn the graph. Second, the semantics of joint type is applied to message passing. The GCN is position-independent and it has no idea of the structural information of skeleton, which is harmful for recognizing actions. We thus take advantage of the semantics of joint type to make the GCN know the structural information during massage passing. 

We denote the type of the $k^{th}$ joint (also referred to as type $k$) by a one-hot vector $\mathbf{j}_{k} \in \mathbb{R}^{d_j}$, where the $k^{th}$ dimension is one and the others are all zeros. Similar to the encoding of position as in Equ.~(\ref{equ:mapping}), we obtain the embedding of the $k^{th}$ joint type as $\widetilde{\mathbf{j}_{k}} \in \mathbb{R}^{C_1}$. 

Given $J$ joints of a skeleton frame, we build a graph of $J$ nodes. We denote the joint representation of joint type $k$ at frame $t$ with both the dynamics and the semantics of joint type as $\textbf{z}_{t,k}$ = [$\textbf{z}_{t,k}, \widetilde{\mathbf{j}_{k}}] \in \mathbb{R}^{2C_{1}}$. All the joints of frame $t$ are then represented by $Z_{t} = (\textbf{z}_{t,1}; \cdots; \textbf{z}_{t,J}) \in \mathbb{R}^{J \times 2C_1}$.

Similar to \cite{wang2018videos,wang2018non, shi2019two}, the edge weight from the $i^{th}$ joint to the $j^{th}$ joint in the same frame $t$ is modeled by their similarity/affinity in the embeded space as
\begin{eqnarray}
    %\begin{aligned}
      \!& S_t(i,j)=  {\theta}(\mathbf{z}_{t,i})^T {\phi}(\mathbf{z}_{t,j}),
    %\end{aligned}
    \label{equ:G}
\end{eqnarray}
where $\theta$ and $\phi$ denote two transformation functions, each implemented by an FC layer, \ie, ${\theta}(\mathbf{x}) = W_3 \mathbf{x} + \mathbf{b}_3 \in \mathbb{R}^{C_2} $ and $ ~{\phi}(\mathbf{x}) = W_4 \mathbf{x} + \mathbf{b}_4 \in \mathbb{R}^{C_2}$. 
%S(\mathbf{z}_{t,i}, \mathbf{z}_{t,j}) =

By computing the affinities of all the joint pairs in the same frame based on (\ref{equ:G}), we obtain the adjacency matrix $S_t \in \mathbb{J\times J}$. Normalization using SoftMax as \cite{vaswani2017attention, wang2018non} is performed on each row of $S_t$ so that the sum of all the edge values connected to a target node is 1. We denote the normalized adjacency matrix by $G_t$. A residual graph convolution layer is used to realize the massage passing among nodes as
\begin{eqnarray}
\label{equ:P}
\begin{aligned}
            \!& Y_t =  G_{t}Z_{t}W_y, \\
            \!& Z'_t = Y_t + Z_tW_z,
\end{aligned}
\end{eqnarray}
where $W_y$ and $W_z$ are transformation matrices. The weight matrices are shared for different temporal frames. $Z'_t$ is the output. Note that one can stack multiple residual graph convolution layers to enable further message passing among nodes with the same adjacency matrix $G_t$.

\subsection{Frame-level Module}

We design a frame-level module to exploit the correlations across frames. To make the network know the order of frames, we incorporate the semantics of frame index to enhance the representation capability of a frame. 
% where joints in the same frame are taken as a whole

We denote the frame index by a one-hot vector $\mathbf{f}_{t} \in \mathbb{R}^{d_{f}}$. Similar to the encoding of position as in Equ.~(\ref{equ:mapping}), we obtain the embedding of the frame index as $\widetilde{\mathbf{f}_{t}} \in \mathbb{R}^{C_{3}}$. We denote the joint representation corresponding to joint type $k$ at frame $t$ with both the semantics of frame index and the learned feature as $\textbf{z}'_{t,k}$ = $\textbf{z}'_{t,k} + \widetilde{\mathbf{f}_{t}} \in \mathbb{R}^{C_{3}}$, where $\textbf{z}'_{t,k} = Z'_t(k,:)$. 
%The frame index is denoted by one hot vector $f_{t} \in \mathbb{R}^{d_{t}}$. We use the same method as Equ. (\ref{equ:mapping}) to obtain the embedding for frame index $\widetilde{\mathbf{f}_{t}} \in \mathbb{R}^{C_{3}}$ and augment the joint representation with the semantics of frame index as $\textbf{z}'_{t,j}$ = $\textbf{z}'_{t,j} + \widetilde{\mathbf{f}_{t}} \in \mathbb{R}^{C_{3}}$. To merge the information of all joints in the same frame, we apply one spatial MaxPooling layer to aggregate them along the spatial dimension. The dimension of the output is $T \times 1 \times C_{3}$. 

To merge the information of all joints in a frame, we apply one spatial MaxPooling layer to aggregate them across the joints. The dimension of feature of the sequence is thus $T\times 1 \times C_{3}$. 
Two CNN layers are applied. The first CNN layer is a temporal convolution layer to model the dependencies of frames. The second CNN layer is used to enhance the representation capability of learned features by mapping it to a high dimension space with kernel size of 1. After the two CNN layers, we apply a temporal MaxPooling layer to aggregate the information of all frames and obtain the sequence level feature representation of $C_{4}$ dimensions. This is then followed by a fully connected layer with Softmax to perform the classification.  %(dimension $C_{4}$).

\section{Experiments}

% In the following, we demonstrate the effectiveness of the proposed semantics-guided neural networks for skeleton-based action recognition. We first describe the datasets and the implementation details in Subsection \ref{dataset}. In Subsection \ref{ablation}, we perform ablation studies to analyze how our model works. In Subsection \ref{compare}, we compare our SGN with the stat-of-the-art approaches on three benchmark datasets. 

%\subsection{Datasets and Implementation Details}
%\label{dataset}
\subsection{Datasets}
\noindent\textbf{NTU60 RGB+D Dataset (NTU60) \cite{shahroudy2016ntu}}. This dataset is collected by the Kinect camera for 3D action recognition with 56,880 skeleton sequences. It contains 60 action classes performed by 40 different subjects. Each human skeleton is represented by 25 joints with 3D coordinates ($J=25$). For the Cross Subject (CS) setting~\cite{shahroudy2016ntu}, half of the 40 subjects are used for training and the rest for testing. For the Cross-View (CV) setting~\cite{shahroudy2016ntu}, the sequences captured by two of the three cameras are used for training and those captured by the other camera are used for testing. Following \cite{shahroudy2016ntu}, we randomly select 10\% of the training sequences for validation for both the CS and CV settings.

\noindent\textbf{NTU120 RGB+D Dataset (NTU120) \cite{liu2019ntu}}. This dataset is an extension of NTU60. It is the largest RGB+D dataset for 3D action recognition with 114,480 skeleton sequences. It contains 120 action classes performed by 106 distinct human subjects. For the Cross Subject (C-Subject) setting, half of the 106 subjects are used for training and the rest for testing. For the Cross Setup (C-Setup) setting, half of the setups are used for training and the rest for testing.

\noindent\textbf{SYSU 3D Human-Object Interaction Dataset (SYSU) \cite{hu2015jointly}}. It contains 480 skeleton sequences of 12 actions performed by 40 different subjects. Each human skeleton has 20 joints ($J=20$). We use the same evaluation protocols as \cite{hu2015jointly}. For the Cross Subject (CS) setting, half of the subjects are used for training and the rest for testing. For the Same Subject (SS) setting, half of the samples of each activity are used for training and the rest for testing. We use the 30-fold cross-validation and show the mean accuracy for each setting \cite{hu2015jointly}.

\subsection{Implementation Details}
\label{implement}
\noindent\textbf{Network Setting}. To obtain the dynamic representation (DR), the number of neurons is set to 64 for each FC layer (\ieno, $C_1=64$). Note that the weights of FC layers are not shared for position and velocity. To encode the joint type, the number of neurons of the two FC layers are both set to 64. To encode the frame index, the numbers of neurons of the two FC layers are set to 64 and 256, respectively and $C_3=256$. For the transformation functions in (\ref{equ:G}), the number of neuron of each FC layer is set to 256, \ieno, $C_2=256$. For the joint-level module, we set the numbers of neurons of the three GCN layers to 128, 256, and 256, respectively. For the fame-level module, we set the number of neurons of the first CNN layer to 256 with kernel size of 3 along the temporal dimension, and set the number of neurons of the second CNN layer to 512 with kernel size of 1 (\ieno, $C_{4}=512$). After each GCN or CNN layer, batch normalization \cite{ioffe2015batch} and ReLU nonlinear activation function are used.

\noindent\textbf{Training}. All experiments are conducted on the Pytorch platform with one P100 GPU card. We use the Adam \cite{kingma2014adam} optimizer with the initial learning rate of 0.001. The learning rate decays by a factor of 10 at the 60$^{th}$ epoch, the 90$^{th}$ epoch, and the 110$^{th}$ epoch, respectively. The training is finished at the 120$^{th}$ epoch. We use a weight decay of 0.0001. The batch sizes for NTU60, NTU120, and SYSU datasets are set to 64, 64 and 16, respectively. Label smoothing \cite{he2019bag} is utilized for all experiments and we set the smoothing factor to 0.1. Cross entropy loss for classification is used to train the networks.

\noindent\textbf{Data Processing}. Similar to \cite{zhang2017view}, sequence level translation based on the first frame is performed to be invariant to the initial positions. If one frame contains two persons, we split the frame into two frames by making each frame contain one human skeleton. During training, according to \cite{liu2016spatio}, we segment the entire skeleton sequence into 20 clips equally, and randomly select one frame from each clip to have a new sequence of 20 frames. During testing, similar to \cite{baradel2018glimpse}, we randomly create 5 new sequences in the similar manner and the mean score is used to predict the class.

During training, we perform data argumentation by randomly rotating the 3D skeletons to some degrees at sequence level to be robust to the view variation. For the NTU60 (CS setting), NTU120, and SYSU datasets, we randomly select three degrees (around $X$, $Y$, $Z$ axes, respectively) between [$-17^\circ, 17^\circ$] for one sequence. Considering that the large view variation for NTU60 (CV setting), we randomly select three degrees between [$-30^\circ , 30^\circ$].

\subsection{Ablation Study}
\label{ablation}

% In Subsection \ref{semantics}, we evaluate the effectiveness of explicitly exploiting semantics. The analysis of effectiveness of hierarchical model is presented in Subsection \ref{hierarchical} The strong baseline and its analyses are introduced in Subsection \ref{strong}. The visualization of the responses of the spatial MaxPooling layer is shown in subsection \ref{vis}. We compare the number of parameters between the proposed SGN and eight other state-of-the-art methods in Subsection \ref{efficiency}

\subsubsection{Effectiveness of Exploiting Semantics}
\label{semantics}

Semantics contains the important structural information of a skeleton sequence which is important for skeleton-based action recognition. To demonstrate the effectiveness of exploiting semantics, by referencing our framework (see Fig.~\ref{fig:framework}), we build eight neural networks and perform various experiments on the NTU60 dataset. Table \ref{tab:sem} shows the comparisons. In the following, \emph{JT} denotes the semantics of joint type, \emph{FI} denotes the semantics of frame index, \emph{G} denotes the learning of graph (adjacency matrix), \emph{P} denotes the graph convolutional operations which enable the massage passing. \emph{T-Conv} denotes the temporal convolutional layer, \ie, the first CNN layer of the frame-level module. Three GCN layers and two CNN layers are used in the joint-level (\textbf{JL}) module and the frame-level (\textbf{FL}) module, respectively. \emph{w} and \emph{w/o}  denote ``with" and ``without", respectively. 

\setlength{\tabcolsep}{5.5pt}
\begin{table}[t]
  \small
  \centering
  \caption{Effectiveness of exploiting semantics in the joint-level module (JL) and frame-level module (FL) on the NTU60 dataset in terms of accuracy (\%). JT denotes joint type and FI denotes frame index.}
    \begin{tabular}{lccc}
    \toprule
    Method  & \#Params(M) & CS & CV \\
    \midrule
    JL(G w/o JT \& P w/o JT) \& FL & 0.62  &86.9 & 92.8 \\
    JL(G w JT \& P w/o JT) \& FL & 0.66    &87.5 & 93.7   \\
    JL(G w/o JT \& P w JT) \& FL & 0.64    &88.6 & 94.1   \\
    JL(G w JT \& P w JT) \& FL & 0.67      &\textbf{88.7} & \textbf{94.1} \\
    \midrule
    JL \& FL(w/o T-Conv) w/o FI &0.54   & 86.8 &92.8 \\
    JL \& FL(w/o T-Conv) w FI   &0.56    &\textbf{87.8} &\textbf{93.7} \\
    \midrule
    JL \& FL(w T-Conv) w/o FI & 0.67   &88.7 &94.1 \\
    JL \& FL(w T-Conv) w FI & 0.69     &\textbf{89.0} &\textbf{94.5} \\
    \bottomrule
    \end{tabular}
  \label{tab:sem}
\end{table}%

\noindent\textbf{Effectiveness of Exploiting Joint Type.}
We investigate four designed models (rows 1 to 4 in Table \ref{tab:sem}) to validate the effectiveness of the joint type on the joint-level module (\textbf{JL}) and all the four models do not include the semantics of temporal index. We explain one model here, and the other three models can be understood in a similar way. ``JL(G w/o JT \& P w/o JT) \& FL" denotes the scheme in which the semantics of joint type is not used for learning graph ($G$) (\ieno, G w/o JT) and does not take part in the graph convolutional operations for massage passing ($P$) (\ieno, P w/o JT) .

%{\color{blue} All expeirental results are shown in Table \ref{tab:sem}.}

We have three main observations as follows.
%about the effectiveness of the semantics of joint type on the joint-level module.

\noindent\textbf{1)} For the learning of graph of skeleton sequence, by introducing the semantics of joint types, ``JL(\textbf{G w JT} \& P w/o JT) \& FL" outperforms ``JL(G w/o JT \& P w/o JT) \& FL" by 0.6\% and 0.9\% for the CS and CV settings, respectively. Intuitively, if the model does not know the types of the joints, it cannot distinguish the joints with the same coordinates even though their semantics are different. The semantics of joint type is beneficial for learning graph edges.

\noindent\textbf{2)} Joint type information is beneficial for message passing in GCN layers. ``JL(G w/o JT \& \textbf{P w JT}) \& FL" is superior to ``JL(G w/o JT \& P w/o JT) \& FL" by 1.7\% and 1.3\% for the CS and CV settings, respectively. The reason is that GCN itself is not aware of the order (type) of joints which makes it hard to learn features of the skeleton data with high structural information. For example, the information contributed from foot joint and wrist joint to a target joint should be different even when the 3D coordinates of the two joints are the same during the message passing. Introducing the joint type information makes GCN more efficient. 

\noindent\textbf{3)} Using the semantics of joint type for both learning graph and the message passing at the same time (``JL(G w JT \& P w JT) \& FL") does not bring further benefits in comparison with ``JL(G w/o JT \& \textbf{P w JT}) \& FL". For message passing $Y_t=G_tZ_tW$ in Equ.~(\ref{equ:P}), the gradient back-propagated to $G_t$ will also be influenced by $Z_t$ which contains joint type information. Actually, $G_t$ is aware of the joint type information implicitly even though we do not include joint type information in the similarity/affinity learning. 

% In summary, the explicit modeling of the joint type information brings benefits to the learning of adjacent matrices and the message passing in the GCN layers.

\noindent\textbf{Effectiveness of Exploiting Frame Index.} We investigate on two models (rows 5 and 6 in Table \ref{tab:sem}) to study the influence of the frame index on the frame-level module (\textbf{FL}) when the temporal convolution is degraded by setting its kernel size to 1. ``JL \& FL(w/o T-Conv) w FI" denotes the model using the semantics of frame index. Both models have incorporated the semantics of joint type. 

Moreover, we investigate two models (rows 7 and 8 in Table \ref{tab:sem}) to study the influence of the frame index when the temporal convolution with kernel size of 3 is used. ``JL \& FL (w T-Conv) w FI" denotes the model using the semantics of frame index. Both models have incorporated the semantics of joint type.

%For the effectiveness of the semantics of temporal index on the frame-level module, there are two observations.
We have two main observations here.

\noindent\textbf{1)} When the temporal convolution is disabled (\ieno, filter kernel size is 1 instead of 3), ``JL \& FL(w/o T-Conv) w \textbf{FI}" outperforms ``JL \& FL(w/o T-Conv) w/o FI" by 1.0\% and 0.9\% for the CS and CV settings, respectively. The frame index information ``tells" the network the frame order of skeleton sequence which is beneficial for action recognition. 

\noindent\textbf{2)} The frame index is helpful for temporal convolution. ``JL \& FL (w T-Conv) w \textbf{FI}" is superior to ``JL \& FL (w T-Conv) w/o FI" by 0.3\% and 0.4\% for the CS and CV settings, respectively. The benefits from the semantics of frame index are smaller than those models without temporal convoluitonal (with filter kernel size of 1). The main reason is the temporal convolutional layer enables the network to know the frame order of skeleton sequence to some extent through large kernel size. However, ``telling" the networks the semantics of frame index explicitly further improves the performance with negligible cost. We take the scheme ``JL \& FL (w T-Conv) w \textbf{FI}" as our final scheme, which is also referred to as ``SGN". 

In summary, the explicit modeling of the joint type information benefits the learning of adjacent matrices and the message passing in the GCN layers. The frame index information enables the model to efficiently exploit the information of sequence order.

%{\color{blue}The frame index information enables the model to efficiently exploit the information of sequence order.}

\subsubsection{Effectiveness of Hierarchical Model}
\label{hierarchical}

We hierarchically model the correlations of the joints in the joint-level module and the frame-level module. To demonstrate its effectiveness, we compare our SGN with two different models and show the results in Table \ref{tab:model}.

%We hierarchically model the correlations of the joints in the joint-level module for modeling the dependencies of joints in the same frame and the frame-level module for modeling the dependencies of frames by taking the joints in the frame as a whole. To demonstrate its effectiveness, we compare SGN with two different models in Table \ref{tab:model}.

``SGN w G-GCN" denotes a non-hierarchical scheme where we remove the spatial MaxPooling layer (SMP), and use the combined semantics (\ieno, joint type and frame index) and dynamics (position and velocity) in the GCN layers. Instead of constructing a graph for each frame, we build a global adaptive graph with all the joints in all the frames and conduct message passing among all those joints. ``SGN w/o SMP" denotes that the spatial MaxPooling layer (SMP) is removed in our scheme ``SGN".

We have the following two observations.
%Table \ref{tab:model} shows the results. %Therefore, the joints in the same frame are not processed as a whole. 
%, which is denoted as G-GCN.

\noindent\textbf{1)} Modeling the correlations of joints of the same frame by GCN is much more effective than modeling the correlations of all joints of all the frames. ``SGN w/o SMP" is superior to ``SGN w G-GCN" by 1.0\% and 0.6\% for the CS and CV settings, respectively. Learning a global content adaptive graph is more complicated and difficult.

\noindent\textbf{2)}  ``SGN" outperforms ``SGN w/o SMP" by 0.7\% and 0.6\% for the CS and CV settings, respectively. Aggregating the information of all joints in a frame by MaxPooing (SMP) plays a role of extracting the representative discriminative information (that has large activation values) of a frame. In addition, the spatial MaxPooling layer reduces the subsequent computation burden.
%The reason is the spatial MaxPooling (across joints within the same frame) is able to select the most informative joints with respect to different actions, which are discussed in Subsection \ref{vis}, to avoid processing the redundancy information in the following layers. 

\setlength{\tabcolsep}{10pt}
\begin{table}[t]
\small
\centering
\caption{Effectiveness of our hierarchical model on the NTU60 dataset in terms of accuracy (\%).}
\begin{tabular}{lccc}
\toprule
Method         & \#Params(M) & CS    & CV    \\
\midrule
SGN w G-GCN      & 0.68     & 87.3 & 93.3  \\
SGN w/o SMP             & 0.69      & 88.3 & 93.9 \\
SGN      & 0.69      & \textbf{89.0} & \textbf{94.5}  \\
\bottomrule
\end{tabular}
\label{tab:model}
\end{table}

\subsubsection{Strong Baseline}
\label{strong}
Previous works usually adopt heavy networks for modeling skeleton sequence of low dimensions \cite{si2018skeleton, si2019attention, shi2019two, zhang2019view}. We exploit some techniques which have been proven very effective in previous works and build a lightweight strong baseline, which has achieved comparable performance as most other state-of-the-art methods \cite{si2018skeleton, zhang2017view, yan2018spatial, gao2019optimized}. We hope this serves as a strong baseline for future research in the skeleton-based action recognition field. All models do not use semantics in this section.

We first build a basic baseline (``Baseline") with the overall pipeline similar to that in Fig.~\ref{fig:framework}. There are three differences. 1) The velocity, joint type, and frame index information are not utilized. 2) Data augmentation (DA) (see Data Processing) is not adopted during training. 3) AveragePooling is used instead of Maxpooling as in \cite{yan2018spatial, shi2019two}. 

Table \ref{tab:skill} shows the influence of our adopted techniques for constructing the strong baseline. We have the following three observations. \textbf{1)} Data augmentation improves the performance significantly for the CV setting. Through the augmentation on the observed views, some ``unseen" views could be ``seen" during the training. \textbf{2)} Two stream networks (using both position and velocity) \cite{si2018skeleton} have proven effective, but two separate networks double the number of parameters. We fuse the two types of information in the early stage (in input) and it improves the performance significantly with only a negligible number of additional parameters (\ieno, 0.01M). \textbf{3)} MaxPooling is much more powerful than AveragePooling. The reason is that MaxPooling works like an attention module which drives to learn and select discriminative features. 

%By jointly adapting these skills, we have a strong baseline with high performance but very small model size.

% The strong baseline has already achieved comparable performance with most other state-of-the-art methods \cite{si2018skeleton, zhang2017view, yan2018spatial, gao2019optimized}. We hope this strong baseline will be useful for other researchers.

\setlength{\tabcolsep}{10pt}
\begin{table}[t]
\small
\centering
\caption{Influence of some techniques on NTU60 dataset in terms of accuracy (\%) and number of parameters.}
\begin{tabular}{lccc}
\toprule
Method         & \#Params(M) & CS    & CV    \\
\midrule
Baseline       & 0.61      & 79.2 & 81.4  \\
+ DA             & 0.61      & 80.6 & 87.1 \\
+ Velocity       & 0.62      & 85.3 & 91.4  \\
+ MaxPooling     & 0.62      & \textbf{86.9} & \textbf{92.8} \\
\bottomrule
\end{tabular}
\label{tab:skill}
\end{table}

\begin{figure}[t]
	\begin{center}
		\includegraphics[width=0.8\linewidth]{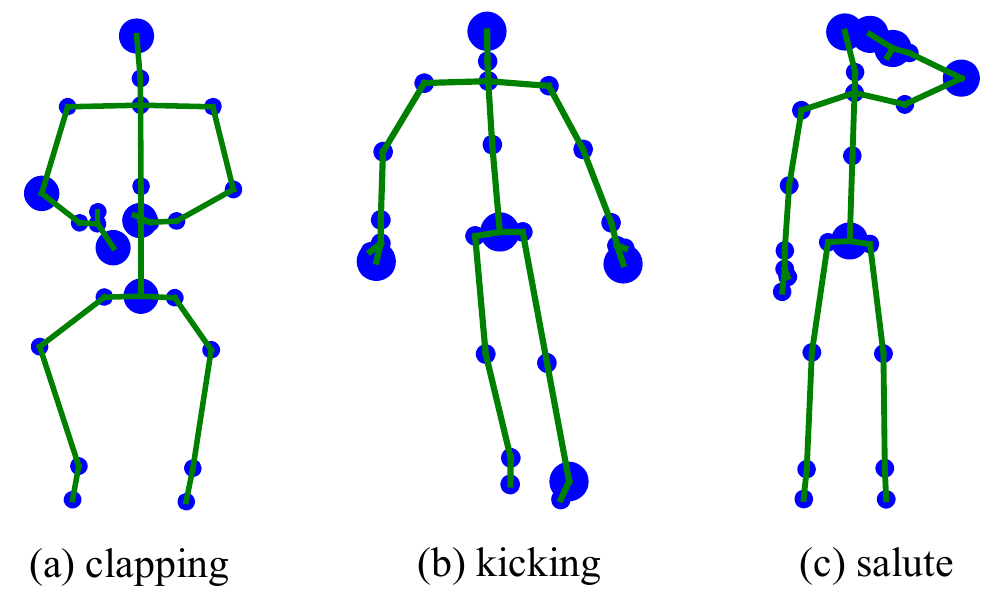}
	\end{center}
	\vspace{-2mm}
	\caption{Visualization of the responses of the spatial MaxPooling layer with respect to three actions, \ie, \emph{clapping}, \emph{kicking}, and \emph{salute}.  The top-5 joints selected by SMP are plotted with larger blue circles.}
	\label{fig:vis}
	%\vspace{-2mm}
\end{figure}

\subsubsection{Visualization of SMP}
\label{vis}

The spatial Maxpooling (SMP) plays a similar role as attention mechanism. We visualize the selected joints by SMP for three actions \ie, \emph{clapping}, \emph{kicking}, and \emph{salute} in Fig.~\ref{fig:vis}. The dimensions of the responses are 256 and each dimension corresponds to one selected joint. We count the times each joint is selected by SMP. The top five chosen joints are shown by large blue circles and the rest are shown by small blue circles. We observe that different actions correspond to different informative joints. The left foot is important for \emph{kicking}. Only the left hand is of great value for \emph{salute}, while both left and right hands are essential for \emph{clapping}. These are consistent with human’s perception.

\subsubsection{Complexity of SGN}
\label{efficiency}

We discuss the complexity of SGN by comparing it with eight state-of-the-art methods for skeleton-based action recognition. As shown in Fig.~\ref{fig:paras}, the number of parameters of VA-RNN \cite{zhang2019view} is the least, but the accuracy is the poorest. VA-CNN\cite{zhang2019view} and 2s-AGCN\cite{shi2019two} achieve good accuracy, but the numbers of parameters are so large. In comparison with the RNN-based, GCN-based, and CNN-based methods, our proposed SGN achieves slightly better performance with much fewer parameters, which makes SGN attractive for many practical applications which have limited computational power.% efficient. The numbers of parameters of those methods range from 0.5M to 24.1M. 

\subsection{Comparison with the State-of-the-arts}
\label{compare}

We compare the proposed SGN with other state-of-the-art methods on the NTU60, NTU 120, and SYSU datasets in Table \ref{tab:ntu}, Table \ref{tab:ntu120}, and Table \ref{tab:SYSU}, respectively. ``SGN w/o Sem." denotes our strong baseline without using semantics.

\setlength{\tabcolsep}{9pt}
\begin{table}[t] %[ht]
   \small
	\centering
	\caption{Performance comparisons on NTU60  with the CS and CV settings in terms of accuracy (\%).}
	\label{tab:ntu}
	\begin{tabular}{lcccc}
		\toprule
		{Method}             &Year                             & CS & CV \\
		\midrule
		HBRNN-L  \cite{du2015hierarchical}          &2015       & 59.1     & 64.0  \\
		Part-Aware LSTM   \cite{shahroudy2016ntu}   &2016            & 62.9     & 70.3  \\
		ST-LSTM + Trust Gate \cite{liu2016spatio}   &2016            & 69.2     & 77.7  \\
		STA-LSTM \cite{song2017end}                 &2017           & 73.4     & 81.2  \\
		GCA-LSTM \cite{liu2017global}               &2017            & 74.4     & 82.8  \\
		Clips+CNN+MTLN \cite{ke2017new}             &2017          & 79.6     & 84.8  \\
		VA-LSTM \cite{zhang2017view}                &2017            & 79.4     & 87.6  \\
		ElAtt-GRU\cite{zhang2018adding}            &2018            & 80.7     & 88.4 \\
		ST-GCN \cite{yan2018spatial}                &2018             & 81.5     & 88.3 \\
		DPRL+GCNN \cite{tang2018deep}               &2018            & 83.5     & 89.8 \\
		SR-TSL  \cite{si2018skeleton}               &2018           & 84.8     & 92.4 \\
		HCN \cite{li2018co}                         &2018           & 86.5     & 91.1 \\
		AGC-LSTM (joint) \cite{si2019attention}      &2019           & 87.5     & 93.5 \\
		AS-GCN \cite{li2019actional}                &2019           & 86.8     & 94.2 \\
		GR-GCN     \cite{gao2019optimized}           &2019          & 87.5     & 94.3 \\          
		2s-AGCN \cite{shi2019two}                   &2019           & 88.5     & \textbf{95.1} \\
		VA-CNN  \cite{zhang2019view}                &2019           & 88.7     & 94.3 \\
		\midrule
		SGN w/o Sem.   &-    & 86.9     &92.8  \\
        SGN             &-   & \textbf{89.0}     & 94.5 \\
		\bottomrule
	\end{tabular}
	\vspace{3mm}
\end{table}

\setlength{\tabcolsep}{3pt}
\begin{table}[t] %[ht]
    \small
	\centering
	\caption{Performance comparisons on  NTU120 with the C-Subject and C-Setup settings in terms of accuracy (\%).}
	\begin{tabular}{lcccc}
		\toprule
		{Method}             &Year                             & C-Subject & C-Setup \\
		\midrule
		Part-Aware LSTM   \cite{shahroudy2016ntu}   &2016            & 25.5    & 26.3  \\
		ST-LSTM + Trust Gate \cite{liu2016spatio}   &2016            & 55.7    & 57.9  \\
		GCA-LSTM \cite{liu2017global}               &2017            & 58.3    & 59.2  \\
		Clips+CNN+MTLN \cite{ke2017new}             &2017            & 58.4    & 57.9  \\
    	Two-Stream GCA-LSTM \cite{liu2017skeleton} &2017             &61.2     &63.3 \\
		RotClips+MTCNN \cite{ke2018learning}        &2018            &62.2     &61.8 \\
		Body Pose Evolution Map \cite{liu2018recognizing} &2018      &64.6     &66.9 \\
		\midrule
		SGN w/o Sem.    &-   &77.4            &79.2 \\
        SGN             &-   &\textbf{79.2}   & \textbf{81.5} \\
		\bottomrule
	\end{tabular}
	\label{tab:ntu120}
	%\vspace{-2mm}
	\vspace{2mm}
\end{table}

As shown in Table \ref{tab:ntu}, the introduction of semantics (\emph{Sem.}) brings performance improvement of 2.1\% and 1.7\% in accuracy for the CS and CV settings, respectively. ``ElAtt-GRU" \cite{zhang2018adding} and ``Clips+CNN+MTLN" \cite{ke2017new} are two representative methods for RNN-based and CNN-based methods, respectively. SGN outperforms them by 8.3\% and 9.4\% in accuracy for the CS setting, respectively. To better explore the structural information of skeleton, some methods \cite{yan2018spatial, si2018skeleton} mix CNN and GCN, or LSTM and GCN together. Our proposed SGN is also superior to \cite{yan2018spatial} and \cite{si2018skeleton} by 5.5\% and 4.2\% in accuracy for the CS setting. The proposed SGN achieves competitive performance when compared to \cite{shi2019two} and \cite{zhang2019view} but with only ten percent of their numbers of parameters as shown in Fig.~\ref{fig:paras}.

As shown in Table \ref{tab:ntu120} and Table \ref{tab:SYSU}, the proposed SGN achieves the best accuracy on NTU120 and SYSU. The NTU120 dataset is a newly released dataset and we compare with the results reported in \cite{liu2019ntu}. Semantics (\emph{sem.}) brings gains of 1.8\% and 2.3\% in accuracy for the C-Subject and the C-Setup settings, respectively. 
%It should be noted that \cite{zhang2018adding} used the pre-trained model on the NTU60 dataset to initialize parameters for the SYSU dataset. The proposed SGN achieves the best performance with or without the pre-trained model.

\section{Conclusion}
In this work, we have presented a simple yet effective end-to-end semantics-guided neural network for high performance skeleton-based human recognition. We explicitly introduce the high level semantics, \ie, joint type and frame index, as part of the network input. To model the correlations of joints, we have proposed a joint-level module for capturing the correlations of joints in the same frame and a frame-level module for modeling the dependencies of frames where all joints in the same frame are taken as a whole. The semantics helps improve the capability of both the GCN and CNN. In addition, we have developed a strong baseline which is better than most previous methods. With an order of magnitude smaller model size than some previous works, our proposed model achieves the state-of-the-art results on three benchmark datasets.

\setlength{\tabcolsep}{9pt}
\begin{table}[t]
   \small
	\centering
	\caption{Performance comparisons on SYSU  in terms of accuracy (\%). * denotes the model uses parameters pre-trained on NTU60.}
	\begin{tabular}{lccc}
		\toprule
		Method  & Year & CS  & SS \\
		\midrule
		VA-LSTM \cite{zhang2017view}       &2017          & 77.5  & 76.9 \\
		ST-LSTM \cite{liu2018skeletontrust}  &2018      & 76.5 & - \\
% 		DPRL+GCNN \cite{tang2018deep}         &2018            & 76.9 &- \\
		GR-GCN     \cite{gao2019optimized}    &2019      & 77.9 & - \\
		Two stream GCA-LSTM  \cite{liu2017skeleton}       &2017       & 78.6 & - \\
		SR-TSL  \cite{si2018skeleton}         &2018      &81.9  & 80.7 \\
		ElAtt-GRU* \cite{zhang2018adding}      &2018          & 85.7 & 85.7 \\
		\midrule
        SGN     & - & 83.0 & 81.6 \\
        SGN*              & -  & \textbf{90.6} & \textbf{89.3} \\
		\bottomrule      
		\label{tab:SYSU}
	\end{tabular}
	%\vspace{-4mm}
\end{table}

\section*{Acknowledgements}
This work was partially supported by the Natural Science Foundation of China (Grant No. 61751308 and 61773311).

{\small
\bibliographystyle{ieee_fullname}
\bibliography{egbib}
}

\end{document}